\documentclass[times]{elsarticle}
\usepackage[top=2.cm,bottom=2.cm,left=2.5cm,right=2.5cm]{geometry}
\usepackage[utf8]{inputenc}
\usepackage{lineno,hyperref}
\usepackage{lscape}
\usepackage{multirow}
\usepackage{amsmath}
\modulolinenumbers[5]
\graphicspath{{figs/}}
\usepackage[table,xcdraw]{xcolor}

\usepackage{orcidlink} 

\usepackage{academicons}

\begin{document}

\begin{frontmatter}

\journal{\dots}
\title{The Evolutionary Computation Methods No One Should Use}
\author{Jakub K\r{u}dela\textsuperscript{\orcidlink{0000-0002-4372-2105}} }
\address{Institute of Automation and Computer Science, Brno University of Technology\\Jakub.Kudela@vutbr.cz}

\begin{abstract}
The center-bias (or zero-bias) operator has recently been identified as one of the problems plaguing the benchmarking
of evolutionary computation methods. This operator lets the methods that utilize it easily optimize functions that have
their respective optima in the center of the feasible set. In this paper, we describe a simple procedure that can be used to
identify methods that incorporate a center-bias operator and use it to investigate 90 evolutionary computation methods
that were published between 1987 and 2022. We show that more than half (47 out of the 90) of the considered
methods have the center-bias problem. We also show that the center-bias is a relatively new phenomenon (with the first
identified method being from 2012), but its inclusion has become extremely prevalent in the last few years. Lastly, we
briefly discuss the possible root causes of this issue.
\end{abstract}


\end{frontmatter}



\section{Introduction}

Imagine the following situation. Encountering a challenging optimization task, you decide to find the most recently developed algorithm for optimization published in some of the most prestigious journals. The analysis of the method performed on several standard benchmarks clearly shows that it is superior to all the other old methods. The paper also contains a link to a repository with the code. So, you give it a try. And it fails. The best results it provides are hardly better (or worse) than the ones you got from a simple implementation of a method that is more than two decades old. Maybe the problem you tried to solve is too challenging? Perhaps a bit of hyperparameter optimization could help the method perform as advertised? Or, maybe the method is not as good as it presented itself.

Through inspiration from natural behaviors, the field of evolutionary computation (EC) produced over its long history a great number of important metaheuristic algorithms, such as Evolutionary Strategy, Genetic Algorithms, Particle Swarm Optimization, or Differential Evolution. Such methods found applications in complex systems where the use of exact algorithms was either inadequate or computationally too prohibitive. However, over the past few years we have witnessed an explosion of "novel" methods that are based on natural/evolutionary principles. The bestiary of EC\footnote{Campelo, F., Aranha, C. Evolutionary computation bestiary. \url{https://github.com/fcampelo/EC-Bestiary}}, which tries to catalog of a portion of these nature-based methods, now contains over 250 methods that claim their inspiration in natural processes. And new methods are emerging at an ever-increasing rate. It is also becoming clearer that there is more creativity being spent at naming these "novel" methods, than in making sure they contain anything new computation-wise. After many of these methods have been found to conceal their lack of novelty behind a methaphor-rich jargon \cite{weyland2010rigorous,camacho2019intelligent,camacho2020grey,villalon2021cuckoo,piotrowski2014novel}, a call was made from within the EC community \cite{aranha2022metaphor}. In the letter, the collective of authors and signatories identified four main issues with the high-volume inflow of new methods: useless metaphors, lack of novelty, poor experimental validation and comparison, and publishing these methods in off-topic journals. 

In this text, we will focus on the poor experimental validation of some of the EC methods. Most of the reasoning about the viability of metaheuristics is done through benchmarking \cite{hellwig2019benchmarking}. If a new method performs well on a universally accepted set of benchmark problems, it is likely to be seen as valid. There have been several benchmark functions/sets proposed over the years, but the most widely recognized ones came from special sessions (competitions) on black-box optimization at two conferences: the IEEE Congress on Evolutionary Computation (CEC), and the Genetic and Evolutionary Computation Conference (GECCO), where the Black-Box Optimization Benchmarking (BBOB) workshop was held.

There is, however, another quite widely used benchmark set that contains some of the most well-known functions such as Griewank, Ackley, Rastrigin, Rosenbrock, and Schwefel. It was recently uncovered \cite{kudela2022critical} that this set contains a serious design flaw, as a large portion of the functions in the set have their respective optimum at a zero vector (or in the centre of the feasible set). This would be fine, if it were not for the methods that utilize this flaw to appear competitive. These methods incorporate a ``check-the-middle'' routine or have a centre-bias (or zero-bias) operator that draws them towards the center of the feasible set. One would expect that such methods do not get published very often, are easily spotted, or at the very least do not appear in high-profile journals.

In this paper, we describe a simple methodology that we use to uncover whether or not a given evolutionary computation method utilizes a center-bias operator. We then investigate 90 evolutionary computation methods from the mealpy library\footnote{N. V. Thieu, “A collection of the state-of-the-art meta-heuristics algorithms in python: Mealpy,” Available: \url{https://doi.org/10.5281/zenodo.3711948}} and Mathworks code repositories\footnote{\url{https://www.mathworks.com/}} for the inclusion of the center-bias. 

\begin{table*}[!t]
\caption{The 13 benchmark functions, dimension 30. U -– unimodal, M –- multimodal, S –- separable, N –- non-separable,  $f^*$ – the optimal function value, $f(0)$ – function value at the zero vector, $x^*$ – optimal solution.}\vspace{2mm}
\label{tab:1}
\centering \small
\resizebox{0.7\linewidth}{!}{
\begin{tabular}{llccrrc}
ID  & name               & \multicolumn{1}{l}{type} & range            & \multicolumn{1}{c}{$f^*$} & \multicolumn{1}{c}{$f(0)$} & \multicolumn{1}{c}{$x*$} \\ \hline 
F01 & Sphere             & U, S                     & {[}-100,100{]}   & 0                      & 0                        & {[}0,0,...{]}          \\
F02 & Schwefel 2.22      & U, N                     & {[}-100,100{]}   & 0                      & 0                        & {[}0,0,...{]}          \\
F03 & Schwefel 1.2       & U, N                     & {[}-100,100{]}   & 0                      & 0                        & {[}0,0,...{]}          \\
F04 & Schwefel 2.21      & U, S                     & {[}-100,100{]}   & 0                      & 0                        & {[}0,0,...{]}          \\
F05 & Rosenbrock         & U, N                     & {[}-30,30{]}     & 0                      & 2.90E+01                 & {[}1,1,...{]}          \\
F06 & Step               & U, S                     & {[}-100,100{]}   & 0                      & 7.50E+00                 & {[}-0.5,-0.5,...{]}    \\
F07 & Quartic with noise & U, S                     & {[}-1.28,1.28{]} & 0                      & 0                        & {[}0,0,...{]}          \\
F08 & Schwefel 2.26      & M, S                     & {[}-500,500{]}   & -1.25E+04              & 0                        & {[}420.9, 420.9,...{]} \\
F09 & Rastrigin          & M, S                     & {[}-5.12,5.12{]} & 0                      & 0                        & {[}0,0,...{]}          \\
F10 & Ackley             & M, N                     & {[}-32,32{]}     & 0                      & 0                        & {[}0,0,...{]}          \\
F11 & Griewank           & M, N                     & {[}-600,600{]}   & 0                      & 0                        & {[}0,0,...{]}          \\
F12 & Penalized1         & M, N                     & {[}-50,50{]}     & 0                      & 1.67E+00                 & {[}-1,-1,...{]}        \\
F13 & Penalized2         & M, S                     & {[}-50,50{]}     & 0                      & 3.00E+00                 & {[}1,1,...{]}         
\end{tabular}
}
\end{table*}

\section{Methodology}
We utilize the same methodology that was used to uncover the center-bias problem in \cite{kudela2022commentary} and \cite{kudela2022critical}. The 13 benchmark function used for our test (and optimization ranges) are shown in Table \ref{tab:1}. One can easily see that many of these functions, apart from F08, have the corresponding optima either at the zero vector or very close to it. The problem F08 is quite different from the rest, as its optimum is far away from the center.

For the evaluation we set the dimension of the problems to 30 and allow for at most 50,000 function evaluations. We also chose a simple performance measure - the mean error (as the difference between the optimal function value and best function value found) over 20 independent runs. Here, we also treat any value smaller than 1e--08 as identical to 1e--08, as the problem is essentially solved and additional precision is not needed (we could treat it as a 0 as well, but we will shortly use fractions of these numbers, which would bring unwanted hassle). We refer to the results of the computation as the ``unshifted'' ones. Afterwards, we introduce a shift operation, that ``moves'' the benchmark function by a predetermined vector $s$, meaning that function $f(x)$ becomes $f(x+s)$. One expects that a ``small'' value of $s$ should not result in a large deviation in the behaviour of the optimization method, as the two problems are very similar. We chose the shift vector as 10\% of the range - e.g., for F01, $s=[20,20,\dots]$. We use the same computational framework (i.e., dimension 30, at maximum 50,000 function evaluations, and 20 independent runs) and refer to the results of these computations as the ``shifted'' ones.

What we are interested in is the ``ratio'' between the ``shifted'' and ``unshifted'' results for the individual benchmark functions, i.e., how many times are the results on the shifted problem worse than on the unshifted one. For the methods that do not incorporate a center-bias, one expects this number to be close to 1 (as the ushifted and shifted problems are similar), while for the methods that include a center-bias, this ratio should be much bigger than 1. Naturally, the value of this ratio will fluctuate depending on the given benchmark function, as well as on the number of independent runs of the algorithms. As a simple indicator of the center-bias, we look at the geometric mean of the ratios for the different benchmarks -- if this value is bigger than 1E+01 (meaning that the method performs roughly at least on order of magnitude better on unshifted problems), we take it as a confirmation of the presence of the center-bias operator.

A small example of this computation is shown in Table \ref{tab:2}, where we investigate five EC methods -- Artificial Bee Colony (ABC) \cite{ABC}, Differential Evolution (DE) \cite{DE}, LSHADE \cite{LSHADE}, Satin Bowerbird Optimizer (SBO) \cite{SBO}, and Runge Kutta Optimizer (RKO) \cite{RKO}. The first two methods (ABC and DE) can be thought of as the ``standard'' ones, LSADE is among the state-of-the ones (as it served as a basis of many of the best methods for recent CEC competitions), and the last two (SBO and RKO) are the ``new'' ones. One can quite easily see that for the first three methods (ABC, DE, and LSHADE) the geometric mean of the ratios is roughly 1, meaning that no center-bias was detected. For SBO, the situation is a bit more complicated, as on many benchmark functions the ratio is relatively low (roughly between 1 and 2), but is very large (almost 5E+04) on F2. This could be a fluke. Fortunately, the nature of the geometric mean will suppress some of the individual flukes - the value for SBO is 3.95E+00 (i.e., $<$1E+01), so we do not label it as a method with a center-bias. The same cannot be said about RKO. Here, many of the ratios are extremely big ($>$1E+06), and the value of the geometric mean is 7.36E+04. We can confidently say that RKO incorporates a center bias operator. 

An interesting observation can be made regarding the benchmark function F08. For all five methods, the ratio between the shifted and unshifted results on F08 is very close to 1. Recall that F08 is the only function in the benchmark set that has the optimum quite far away from the center of the feasible set, and its function value at the zero-vector is also quite far away from the optimal value. Although it is arguably not surprising that the methods have a ratio around 1 on this function, it is still valuable to have it confirmed -- the function F08 serves as a sanity check in the benchmark set.

\begin{table}[!t]
\caption{The results of proposed methodology demonstrated on five methods.}
\label{tab:2}
\resizebox{1\linewidth}{!}{
\begin{tabular}{l|rrr|rrr|rrr|rrr|rrr}
        & \multicolumn{1}{l}{}         & \multicolumn{1}{l}{ABC}     & \multicolumn{1}{l|}{}      & \multicolumn{1}{l}{}         & \multicolumn{1}{l}{DE}      & \multicolumn{1}{l|}{}      & \multicolumn{1}{l}{}          & \multicolumn{1}{l}{LSHADE}  & \multicolumn{1}{l|}{}      & \multicolumn{1}{l}{}         & \multicolumn{1}{l}{SBO}     & \multicolumn{1}{l|}{}      & \multicolumn{1}{l}{}         & \multicolumn{1}{l}{RKO}     & \multicolumn{1}{l}{}      \\
        & \multicolumn{1}{c}{unshifted} & \multicolumn{1}{c}{shifted} & \multicolumn{1}{c|}{ratio} & \multicolumn{1}{c}{unshifted} & \multicolumn{1}{c}{shifted} & \multicolumn{1}{c|}{ratio} & \multicolumn{1}{c}{unshifted} & \multicolumn{1}{c}{shifted} & \multicolumn{1}{c|}{ratio} & \multicolumn{1}{c}{unshifted} & \multicolumn{1}{c}{shifted} & \multicolumn{1}{c|}{ratio} & \multicolumn{1}{c}{unshifted} & \multicolumn{1}{c}{shifted} & \multicolumn{1}{c}{ratio} \\ \hline 
F1  & 5.19E+03 & 4.19E+03 & 8.08E$-$01 & 3.82E$-$02 & 2.61E$-$02 & 6.83E$-$01 & 1.00E$-$08 & 1.00E$-$08 & 1.00E+00 & 5.85E+00 & 1.20E+01 & 2.05E+00 & 1.00E$-$08 & 1.30E$-$06 & 1.30E+02 \\
F2  & 3.08E+03 & 5.95E+02 & 1.93E$-$01 & 2.79E+00 & 3.06E+00 & 1.10E+00 & 6.37E$-$06 & 9.55E$-$06 & 1.50E+00 & 1.25E+15 & 6.19E+19 & 4.95E+04 & 1.00E$-$08 & 2.05E+00 & 2.05E+08 \\
F3  & 8.16E+04 & 8.04E+04 & 9.86E$-$01 & 1.89E+04 & 1.64E+04 & 8.68E$-$01 & 4.13E$-$04 & 4.16E$-$04 & 1.01E+00 & 1.48E+04 & 9.08E+04 & 6.14E+00 & 1.00E$-$08 & 5.03E$-$02 & 5.03E+06 \\
F4  & 5.81E+01 & 5.82E+01 & 1.00E+00 & 1.35E+01 & 1.31E+01 & 9.70E$-$01 & 5.08E$-$03 & 3.75E$-$03 & 7.38E$-$01 & 1.28E+01 & 1.98E+01 & 1.55E+00 & 1.00E$-$08 & 2.54E+00 & 2.54E+08 \\
F5  & 1.40E+03 & 1.38E+04 & 9.86E+00 & 6.09E+02 & 5.88E+02 & 9.66E$-$01 & 2.07E+01 & 2.12E+01 & 1.02E+00 & 5.63E+02 & 1.17E+03 & 2.07E+00 & 2.53E+01 & 3.99E+01 & 1.58E+00 \\
F6  & 5.36E+03 & 6.12E+03 & 1.14E+00 & 6.68E+00 & 7.10E+00 & 1.06E+00 & 1.00E$-$08 & 1.00E$-$08 & 1.00E+00 & 1.35E+01 & 2.42E+01 & 1.79E+00 & 1.00E$-$08 & 1.74E$-$06 & 1.74E+02 \\
F7  & 1.45E+00 & 2.26E+00 & 1.55E+00 & 8.56E$-$02 & 8.18E$-$02 & 9.56E$-$01 & 3.19E$-$03 & 3.33E$-$03 & 1.04E+00 & 4.37E$-$01 & 8.21E$-$01 & 1.88E+00 & 9.26E$-$05 & 1.64E$-$02 & 1.76E+02 \\
F8  & 5.07E+03 & 5.79E+03 & 1.14E+00 & 1.43E+04 & 1.50E+04 & 1.05E+00 & 1.64E+02 & 1.80E+02 & 1.10E+00 & 6.99E+03 & 5.99E+03 & 8.58E$-$01 & 4.19E+03 & 4.90E+03 & 1.17E+00 \\
F9  & 2.49E+01 & 3.03E+01 & 1.22E+00 & 3.98E+02 & 3.98E+02 & 1.00E+00 & 1.08E+01 & 1.10E+01 & 1.02E+00 & 3.66E+01 & 5.16E+01 & 1.41E+00 & 1.00E$-$08 & 2.99E+01 & 2.99E+09 \\
F10 & 8.69E+00 & 7.57E+00 & 8.70E$-$01 & 1.53E+00 & 1.44E+00 & 9.41E$-$01 & 2.38E$-$07 & 2.57E$-$07 & 1.08E+00 & 4.72E+00 & 5.48E+00 & 1.16E+00 & 1.00E$-$08 & 2.86E+00 & 2.86E+08 \\
F11 & 3.16E+02 & 3.35E+02 & 1.06E+00 & 1.07E+00 & 1.06E+00 & 9.91E$-$01 & 1.00E$-$08 & 1.00E$-$08 & 1.00E+00 & 1.06E+00 & 1.10E+00 & 1.04E+00 & 1.00E$-$08 & 1.25E$-$02 & 1.25E+06 \\
F12 & 7.71E+05 & 1.09E+06 & 1.41E+00 & 1.13E+00 & 1.27E+00 & 1.12E+00 & 1.00E$-$08 & 1.00E$-$08 & 1.00E+00 & 6.81E+00 & 1.29E+01 & 1.89E+00 & 1.00E$-$08 & 3.37E+00 & 3.37E+08 \\
F13 & 9.98E+05 & 5.45E+06 & 5.46E+00 & 6.36E+00 & 5.94E+00 & 9.34E$-$01 & 1.00E$-$08 & 1.00E$-$08 & 1.00E+00 & 1.51E+00 & 4.63E+00 & 3.07E+00 & 4.40E$-$03 & 1.18E$-$02 & 2.68E+00 \\ \hline 
geomean & \multicolumn{1}{l}{-}        & \multicolumn{1}{l}{-}       & 1.29E+00                  & \multicolumn{1}{l}{-}        & \multicolumn{1}{l}{-}       & 9.66E$-$01                  & \multicolumn{1}{l}{-}         & \multicolumn{1}{l}{-}       & 1.03E+00                  & \multicolumn{1}{l}{-}        & \multicolumn{1}{l}{-}       & 3.95E+00                  & \multicolumn{1}{l}{-}        & \multicolumn{1}{l}{-}       & 7.36E+04                 
\end{tabular}
}
\end{table}

\section{Results and Discussion}
\begin{figure}[!h]
    \centering
    \includegraphics[width = .95\linewidth]{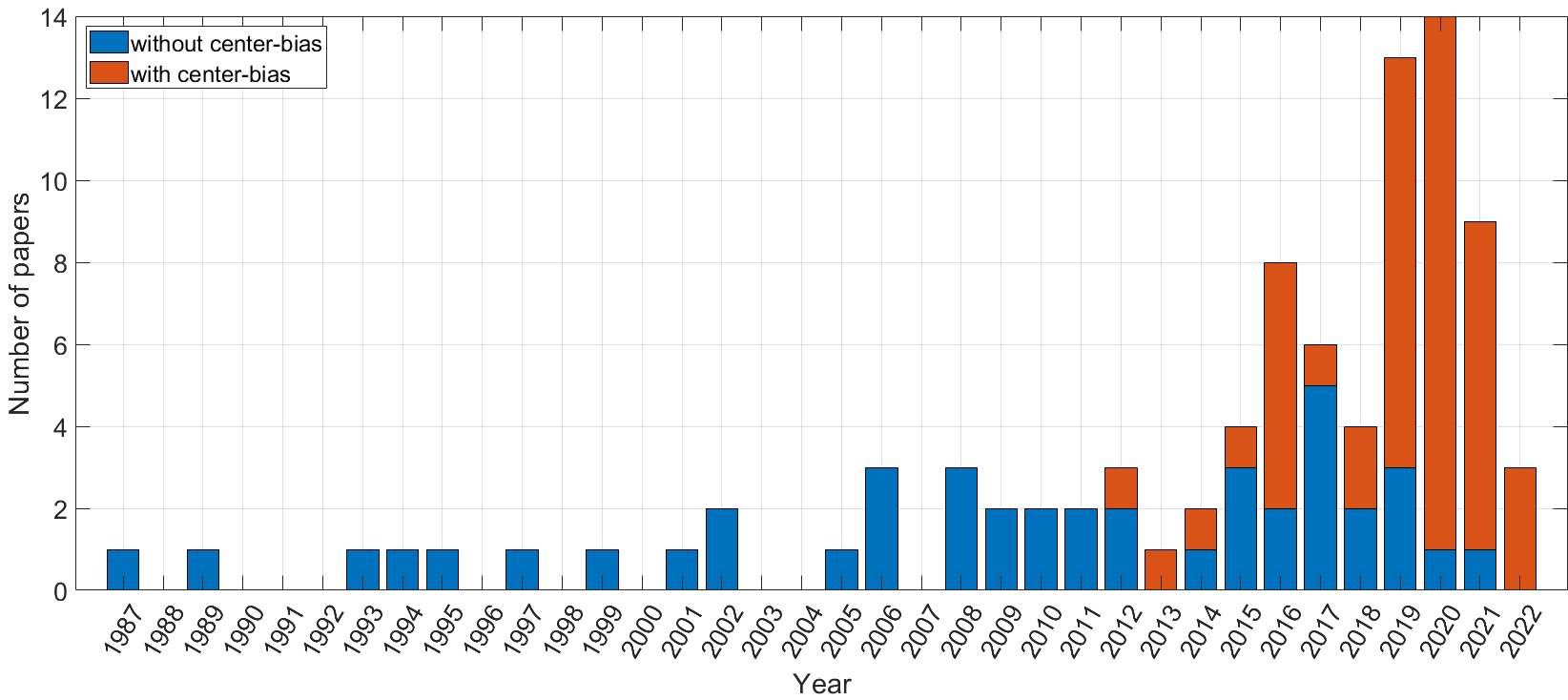}\vspace{-2mm}
    \caption{Number of papers proposing methods with/without center-bias in time.}
    \label{f:graph}
\end{figure}

In this section we report the results of using the methodology described in the previous section on 90 selected EC methods. The selected methods, the year of the publication that describes them, and the geometric mean of the ratios are shown (in alphabetical order) in Table \ref{tab:3}, with the ones with a confirmed center-bias (i.e., values $>$1E+01) highlighted in red. These results are extremely worrying, as more than a half (47 out of the 90) methods have a confirmed center-bias. And they become even worse when we take a look at the number of methods with center-bias that were proposed recently, as shown in Figure \ref{f:graph}.

\newpage

\begin{table}[!t]
\centering
\label{tab:3}
\caption{Considered algorithms and results.}
\resizebox{0.99\linewidth}{!}{.

\begin{tabular}{llrr|llrr}
Abbreviation & Method name & Year & Geomean & Abbreviation & Method name & Year & geomean \\ \hline 
ABC \cite{ABC}    & Artificial Bee Colony                     & 2008 & 1.29E+00                         & HC \cite{HC}      & Hill Climbing                        & 1993 & 1.13E+00                         \\
ACOR \cite{ACOR}   & Ant Colony Optimization Continuous        & 2008 & 7.40E-01                         & HGS \cite{HGS}     & Hunger Games Search                  & 2021 & \cellcolor[HTML]{F4C7C3}3.69E+06 \\
AEO \cite{AEO}   & Artificial Ecosystem-based Optimization   & 2020 & \cellcolor[HTML]{F4C7C3}1.01E+07 & HGSO \cite{HGSO}    & Henry Gas Solubility Optimization    & 2019 & \cellcolor[HTML]{F4C7C3}8.07E+03 \\
ALO \cite{ALO}   & Ant Lion Optimizer                        & 2015 & 1.44E+00                         & HHO \cite{HHO}      & Harris Hawks Optimization            & 2019 & \cellcolor[HTML]{F4C7C3}1.62E+05 \\
AO \cite{AO}    & Aquila Optimization                       & 2021 & \cellcolor[HTML]{F4C7C3}2.26E+05 & HS \cite{HS}       & Harmony Search                       & 2001 & 9.97E-01                         \\
AOA \cite{AOA}   & Arithmetic Optimization Algorithm         & 2021 & \cellcolor[HTML]{F4C7C3}1.01E+10 & IWO \cite{IWO}      & Invasive Weed Optimization           & 2006 & 1.88E+00                         \\
ArchOA \cite{ArchOA}& Archimedes Optimization Algorithm         & 2021 & \cellcolor[HTML]{F4C7C3}3.75E+07 & JA \cite{JA}       & Jaya Algorithm                       & 2016 & \cellcolor[HTML]{F4C7C3}1.19E+01 \\
ASO \cite{ASO}   & Atom Search Optimization                  & 2019 & 8.71E-01                         & KMA \cite{KMA}      & Komodo Mlipir Algorithm              & 2022 & \cellcolor[HTML]{F4C7C3}1.84E+05 \\
BA \cite{BA}    & Bat-inspired Algorithm                    & 2010 & 1.44E+00                         & LCO \cite{LCO}      & Life Choice-based Optimization       & 2020 & \cellcolor[HTML]{F4C7C3}8.31E+07 \\
BBO \cite{BBO}   & Biogeography-Based Optimization           & 2008 & 6.43E-01                         & MA \cite{MA}       & Memetic Algorithm                    & 1989 & 1.68E-03                         \\
BeesA \cite{BeesA} & Bees Algorithm                            & 2006 & 1.16E+00                         & MFO \cite{MFO}      & Moth-Flame Optimization              & 2015 & 1.73E-01                         \\
BES \cite{BES}   & Bald Eagle Search                         & 2020 & \cellcolor[HTML]{F4C7C3}2.62E+08 & MGO \cite{MGO}     & Mountain Gazelle Optimizer           & 2022 & \cellcolor[HTML]{F4C7C3}1.28E+01 \\
BFO \cite{BFO}    & Bacterial Foraging Optimization           & 2002 & 9.66E-01                         & MPA \cite{MPA}      & Marine Predators Algorithm           & 2020 & \cellcolor[HTML]{F4C7C3}1.06E+02 \\
BOA \cite{BOA}    & Butterfly Optimization Algorithm          & 2019 & \cellcolor[HTML]{F4C7C3}9.57E+05 & MRFO \cite{MRFO}     & Manta Ray Foraging Optimization      & 2020 & \cellcolor[HTML]{F4C7C3}6.40E+07 \\
BRO \cite{BRO}    & Battle Royale Optimization                & 2021 & \cellcolor[HTML]{F4C7C3}2.59E+09 & MSA \cite{MSA}      & Moth Search Algorithm                & 2018 & 8.37E+00                         \\
BSA \cite{BSA}    & Bird Swarm Algorithm                      & 2016 & \cellcolor[HTML]{F4C7C3}1.09E+01 & MVO \cite{MVO}      & Multi-Verse Optimizer                & 2016 & 1.75E+00                         \\
BSO \cite{BSO}    & Brain Storm Optimization                  & 2011 & 7.85E+00                         & NMRA \cite{NMRA}     & Naked Mole-Rat Algorithm             & 2019 & \cellcolor[HTML]{F4C7C3}5.65E+08 \\
CA \cite{CA}     & Culture Algorithm                         & 2009 & 7.18E-01                         & NRO \cite{NRO}      & Nuclear Reaction Optimization        & 2019 & \cellcolor[HTML]{F4C7C3}2.30E+06 \\
CEM \cite{CEM}    & Cross-Entropy Method                      & 2005 & 1.33E+00                         & PFA \cite{PFA}      & Pathfinder Algorithm                 & 2019 & \cellcolor[HTML]{F4C7C3}3.11E+08 \\
CGO \cite{CGO}    & Chaos Game Optimization                   & 2021 & \cellcolor[HTML]{F4C7C3}2.14E+07 & PSO \cite{PSO}      & Particle Swarm Optimization          & 1995 & 9.70E-01                         \\
ChOA \cite{ChOA}    & Chimp optimization algorithm              & 2020 & \cellcolor[HTML]{F4C7C3}3.89E+03 & PSS \cite{PSS}      & Pareto-like Sequential Sampling      & 2021 & \cellcolor[HTML]{F4C7C3}1.77E+03 \\
COA  \cite{COA}  & Coyote Optimization Algorithm             & 2018 & \cellcolor[HTML]{F4C7C3}4.00E+06 & QSA \cite{QSA}     & Queuing Search Algorithm             & 2021 & 7.91E-01                         \\
CRO \cite{CRO}    & Coral Reefs Optimization                  & 2014 & 9.69E-01                         & RKO \cite{RKO}      & Runge Kutta Optimizer                & 2021 & \cellcolor[HTML]{F4C7C3}7.36E+04 \\
CSA \cite{CSA}    & Cuckoo Search Algorithm                   & 2009 & 1.10E+00                         & SA \cite{SA}       & Simulated Annealing                  & 1987 & 8.95E-01                         \\
CSO \cite{CSO}    & Cat Swarm Optimization                    & 2006 & 9.58E-01                         & SARO \cite{SARO}     & Search And Rescue Optimization       & 2019 & 2.27E+00                         \\
DE \cite{DE}     & Differential Evolution                    & 1997 & 9.66E-01                         & SBO \cite{SBO}      & Satin Bowerbird Optimizer            & 2017 & 3.95E+00                         \\
DandO \cite{DandO}  & Dandelion Optimizer                       & 2022 & \cellcolor[HTML]{F4C7C3}3.59E+02 & SCA \cite{SCA}      & Sine Cosine Algorithm                & 2016 & \cellcolor[HTML]{F4C7C3}1.18E+04 \\
DO \cite{DO}     & Dragonfly Optimization                    & 2016 & \cellcolor[HTML]{F4C7C3}6.62E+02 & SFO \cite{SFO}      & SailFish Optimizer                   & 2019 & \cellcolor[HTML]{F4C7C3}2.57E+07 \\
EFO \cite{EFO}    & Electromagnetic Field Optimization        & 2016 & 6.78E-01                         & SHO \cite{SHO}      & Spotted Hyena Optimizer              & 2017 & 1.31E+00                         \\
EHO \cite{EHO}    & Elephant Herding Optimization             & 2015 & \cellcolor[HTML]{F4C7C3}3.99E+03 & SLO \cite{SLO}      & Sea Lion Optimization Algorithm      & 2019 & 2.83E+00                         \\
EO \cite{EO}     & Equilibrium Optimizer                     & 2020 & \cellcolor[HTML]{F4C7C3}4.65E+03 & SMA \cite{SMA}      & Slime Mould Algorithm                & 2020 & \cellcolor[HTML]{F4C7C3}4.54E+06 \\
EOA \cite{EOA}    & Earthworm Optimisation Algorithm          & 2018 & \cellcolor[HTML]{F4C7C3}2.55E+05 & SRSR \cite{SRSR}     & Swarm Robotics Search And Rescue     & 2017 & 2.03E+00                         \\
EP \cite{EP}     & Evolutionary Programming                  & 1999 & 1.43E+00                         & SSA \cite{SSA}      & Sparrow Search Algorithm             & 2020 & \cellcolor[HTML]{F4C7C3}2.61E+06 \\
ES \cite{ES}     & Evolution Strategies                      & 2002 & 1.14E+00                         & SSDO \cite{SSDO}     & Social Ski-Driver Optimization       & 2020 & \cellcolor[HTML]{F4C7C3}5.40E+08 \\
FA \cite{FA}     & Fireworks Algorithm                       & 2010 & 1.34E+00                         & SSO \cite{SSO}      & Salp Swarm Optimization              & 2017 & \cellcolor[HTML]{F4C7C3}2.28E+01 \\
FBIO \cite{FBIO}   & Forensic-Based Investigation Optimization & 2020 & \cellcolor[HTML]{F4C7C3}5.07E+06 & SSpiderA \cite{SSpiderA} & Social Spider Algorithm              & 2015 & 1.12E+00                         \\
FFA \cite{FFA}    & Firefly Algorithm                         & 2011 & 1.18E+00                         & STOA \cite{STOA}     & Sooty Tern Optimization Algorithm    & 2019 & \cellcolor[HTML]{F4C7C3}6.78E+04 \\
FOA \cite{FOA}    & Fruit-fly Optimization Algorithm          & 2012 & 4.01E+00                         & TLO \cite{TLO}      & Teaching Learning-based Optimization & 2012 & \cellcolor[HTML]{F4C7C3}3.19E+04 \\
FPA \cite{FPA}    & Flower Pollination Algorithm              & 2012 & 9.74E-01                         & TPO \cite{TPO}      & Tree Physiology Optimization         & 2019 & \cellcolor[HTML]{F4C7C3}2.20E+01 \\
GA \cite{GA}     & Genetic Algorithm                         & 1994 & 1.02E+00                         & TSA \cite{TSA}      & Tunicate Swarm Algorithm             & 2020 & \cellcolor[HTML]{F4C7C3}6.25E+06 \\
GBO \cite{GBO}    & Gradient-Based Optimizer                  & 2020 & \cellcolor[HTML]{F4C7C3}7.17E+07 & TWO \cite{TWO}      & Tug of War Optimization              & 2017 & 9.69E-01                         \\
GCO \cite{GCO}    & Germinal Center Optimization              & 2018 & 1.02E+00                         & VCS \cite{VCS}      & Virus Colony Search                  & 2016 & \cellcolor[HTML]{F4C7C3}2.90E+04 \\
GOA \cite{GOA}    & Grasshopper Optimization Algorithm        & 2017 & 3.39E+00                         & WDO \cite{WDO}      & Wind Driven Optimization             & 2013 & \cellcolor[HTML]{F4C7C3}4.86E+01 \\
GSKA \cite{GSKA}   & Gaining Sharing Knowledge-based Algorithm & 2020 & 4.51E-01                         & WHO \cite{WHO}      & Wildebeest Herd Optimization         & 2019 & \cellcolor[HTML]{F4C7C3}8.63E+02 \\
GWO \cite{GWO}    & Grey Wolf Optimizer                       & 2014 & \cellcolor[HTML]{F4C7C3}8.89E+05 & WOA \cite{WOA}      & Whale Optimization Algorithm         & 2016 & \cellcolor[HTML]{F4C7C3}1.87E+03
\end{tabular}
}
\end{table}

We can find that while the number newly proposed methods that do not have the center-bias problem increased only slightly over the last three decades, the number of methods that we have identified as having a center-bias problem is growing extremely fast, especially in the last five years. It has gotten so bad that an overwhelming majority of newly proposed methods have the center-bias problem. An important thing to remark is that we only considered the ``baseline'' (or original) versions of the methods, and not any of the ``improved'' or ``enhanced'' variants that are also being published at an ever-increasing rate. If these were considered as well, we suspect that the graph would look even worse.

We can also see that the first method that we have found to incorporate the center-bias was Teaching Learning-based Optimization (TLO) in 2012, followed by Wind Driven Optimization (WDO) in 2013, and Grey Wolf Optimizer (GWO) in 2014. From these three, TLO and GWO have become extremely influential (gathering thousand of citations) and spawned a large number of variants and modifications. Our failure to quickly identify that they are defective is one of the root causes of the mess we have to deal with now. Although the defect of the GWO was uncovered in 2019 \cite{niu2019defect}, GWO is still used in numerical comparisons (even on problems that are susceptible to the center-bias operator). Similar defects have been also found for the Salp Swarm Optimization (SSO), Sooty Tern Optimization Algorithm (STOA), Tunicate Swarm Algorithm (TSA), Harris Hawks Optimization (HHO), Butterfly Optimization Algorithm (BOA), Slime Mould Algorithm (SMA), Gradient-Based Optimizer (GBO), Marine Predators Algorithm (MPA), and Komodo Mlipir Algorithm (KMA), all in 2022 \cite{castelli2022salp,kudela2022commentary,kudela2022critical}.

For the most part, the methods that incorporate a center-bias procedure have been developed by a diverse groups of authors (i.e., most authors have only one or two such methods). There is, however, one very notable exception. The group of S. Mirjalili, A. H. Gandomi, and A. A. Heidari is collectively responsible for 20 of the 47 methods that contain center-bias (and S. Mirjalili is also one of the authors of GWO).

Another interesing point to make is that some of methods that display the worst center-bias properties (i.e., the largest values of the geometric mean of the ratios) are the ones which were supposedly based on ``mathematical'' processes -- Arithmetic Optimization Algorithm (AOA), Gradient-Based Optimizer (GBO), Runge Kutta Optimizer (RKO), and Sine Cosine Algorithm (SCA). The following are the first few sentences from the abstract of the paper describing RKO \cite{RKO}:\vspace{1mm}

\noindent\fbox{%
    \parbox{\textwidth}{%
       ``The optimization field suffers from the metaphor-based “pseudo-novel” or “fancy” optimizers. Most of these cliché methods mimic animals' searching trends and possess a small contribution to the optimization process itself. Most of these cliché methods suffer from the locally efficient performance, biased verification methods on easy problems, and high similarity between their components' interactions. This study attempts to go beyond the traps of metaphors and introduce a novel metaphor-free population-based optimization method based on the mathematical foundations and ideas of the Runge Kutta (RK) method widely well-known in mathematics.''
    }%
}\vspace{1mm}
The irony is rich.

\section{Conclusion} \label{S:5}
The center-bias problem is right now one of the major issues plaguing the field of evolutionary computation. In this paper, we have described a simple procedure for identifying methods with center-bias and used it to investigate 90 methods that were proposed in the last three decades. We have found that 47 of the 90 methods utilize center-bias. We have also shown that the utilization of center-bias is a relatively new phenomenon, with first instances from 2012-2014. However, the number of methods that use it grew extremely fast in the last five years.

We should note that there is an additional problem that plagues the field right now, which is the equivalence of some of the methods that is hidden under a metaphore-rich jargon. Some of the methods that we have identified as not having a center bias, such as Harmony Search (HS), Cockoo Search Algorith (CSA), Firefly Algorithm (FA), Moth-Flame Optimization (MFO), Ant Lion Optimizer (ALO) should also not be used, as they have been found to be either extremely similar (or identical) to other methods \cite{weyland2010rigorous,villalon2021cuckoo,camacho2022exposing}.

Further utilization, development and improvement of the methods that contain a center-bias is an exercise in futility, as by their very nature they cannot be considered as efficient algorithms. Enough computational and human resources were already wasted in writing, testing, comparing, and reviewing these methods. The field of evolutionary computation needs a spring cleaning. The sooner the better.

\section*{Acknowledgment}
This work was supported by IGA BUT: FSI-S-20-6538.

\bibliography{bib.bib}

%






\end{document}